\documentclass[letterpaper]{article} 
\usepackage{aaai2026}  
\usepackage{times}  
\usepackage{helvet}  
\usepackage{courier}  
\usepackage[hyphens]{url}  
\usepackage{graphicx} 
\usepackage{natbib}  
\usepackage{caption} 
\usepackage{algorithm}
\usepackage{algorithmic}
\usepackage{multirow}
\usepackage{booktabs}
\usepackage{amssymb}
\usepackage{amsmath}
\usepackage{marvosym}
\usepackage{float}
\usepackage{newfloat}
\usepackage{listings}
\DeclareCaptionStyle{ruled}{labelfont=normalfont,labelsep=colon,strut=off} 
\floatstyle{ruled}
\newfloat{listing}{tb}{lst}{}
\floatname{listing}{Listing}
\title{OmniPT: Unleashing the Potential of Large Vision Language Models for Pedestrian Tracking and Understanding}
\author{
    Teng Fu,
    Mengyang Zhao,
    Ke Niu,
    Kaixin Peng,
    Bin Li\textsuperscript{\Letter} 
}
\affiliations{
    Shanghai Key Laboratory of Intelligent Information Processing\\ College of Computer Science and Artificial Intelligence, Fudan University\\
    \{tfu23, kniu22, kxpeng25\}@m.fudan.edu.cn, \{myzhao20, libin\}@fudan.edu.cn
}

\begin{document}
\maketitle
\renewcommand{\thefootnote}{}
\footnotetext{\Letter ~Corresponding author.}
\begin{figure*}[h] 
    \centering
    \includegraphics[width=.95\textwidth]{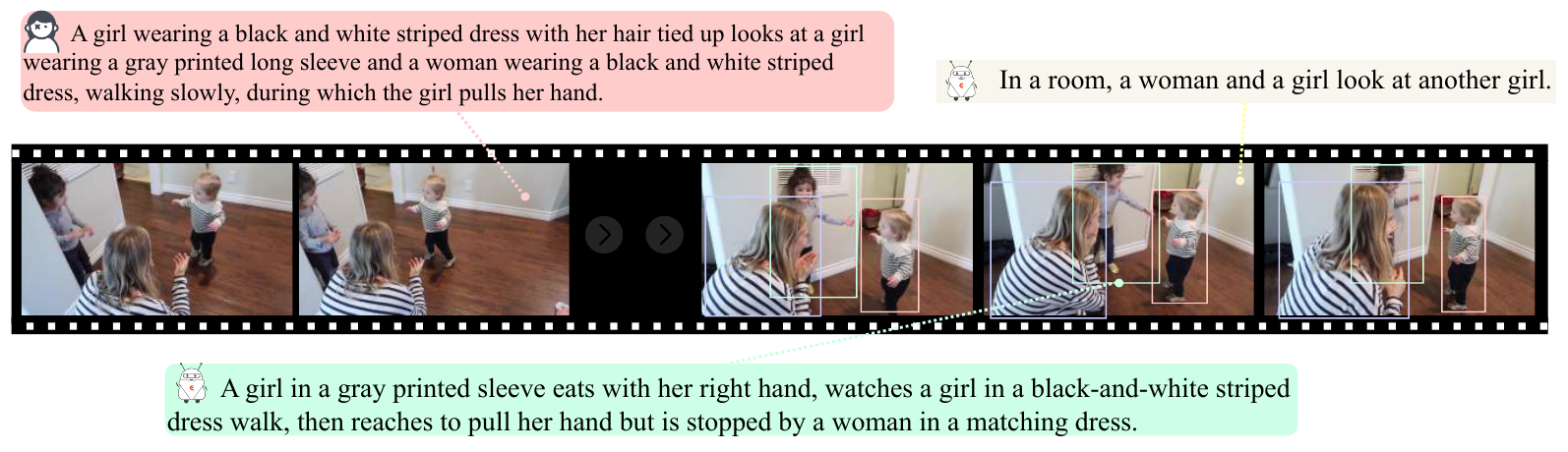}
    \caption{OmniPT can perform traditional object tracking; it can also perform object-specific tracking based on given references and can perform semantic understanding of the tracked objects as well as the whole video.}
    \label{overview}
\end{figure*}

\begin{abstract}
LVLMs have been shown to perform excellently in image-level tasks such as VQA and caption. However, in many instance-level tasks, such as visual grounding and object detection, LVLMs still show performance gaps compared to previous expert models. Meanwhile, although pedestrian tracking is a classical task, there have been a number of new topics in combining object tracking and natural language, such as Referring MOT, Cross-view Referring MOT, and Semantic MOT. These tasks emphasize that models should understand the tracked object at an advanced semantic level, which is exactly where LVLMs excel. In this paper, we propose a new unified \textbf{P}edestrian \textbf{T}racking framework, namely \textbf{OmniPT}, which can track, track based on reference and generate semantic understanding of tracked objects interactively. We address two issues: how to model the tracking task into a task that foundation models can perform, and how to make the model output formatted answers. To this end, we implement a training phase consisting of \texttt{RL-Mid Training-SFT-RL}. Based on the pre-trained weights of the LVLM, we first perform a simple RL phase to enable the model to output fixed and supervisable bounding box format. Subsequently, we conduct a mid-training phase using a large number of pedestrian-related datasets. Finally, we perform supervised fine-tuning on several pedestrian tracking datasets, and then carry out another RL phase to improve the model's tracking performance and enhance its ability to follow instructions. We conduct  experiments on tracking benchmarks and the experimental results demonstrate that the proposed method can perform better than the previous methods.
\end{abstract}

\section{Introduction}
\label{sec:intro}

Pedestrian Tracking~\cite{fu2023denoising, cai2022memot, zeng2022motr, zhang2023motrv2} is a classical task in computer vision, which aims to locate each person in a sequence of images and assign a unique id to each object. In recent years, the task has had a wide range of applications, such as autonomous driving, intelligent surveillance, and sports analytics. However, pedestrian tracking in complex scenarios is still a challenge, and how to stably track an object when it is frequently obscured, blurred, or even disappeared is still a popular and practical topic.

Instead, humans can easily keep tracking an object, even if that object disappears after a long period of time. In medicine, this is a joint result of the semantic and situational memory functions in humans~\cite{martin2001semantic, binder2011neurobiology, kumar2021semantic}. Simply put, we will abstract an object into a semantic description so that we can subsequently perform a subconscious retrieval to recognize the object. 
This provides new promising ideas for solving the above difficulties that still exist in MOT topics: \textbf{Stable semantic information} about the object can be extracted as special appearance features in the traditional sense to assist the tracker in tracking. At the same time, many new multimodal topics have emerged in the MOT field, \textit{i. e.}, tracking based on verbal information (named Referring MOT, RMOT~\cite{wu2023referring, zhang2024bootstrapping, fu2025crowdtrack}); cross-view object tracking that tracks based on verbal information (named Cross View Referring MOT, CRMOT~\cite{chen2024cross}) and semantic multiple object tracking for semantic understanding of tracked objects and videos (named Semantic MOT, SMOT~\cite{li2025beyond}). These tasks all emphasize that while tracking an object, the model should understand the object at the semantic level.

Meanwhile, LVLMs have demonstrated remarkable performance in image-level understanding, but their capabilities at the instance level and pixel level still have certain gaps compared with previous expert models. In this paper, we propose \textbf{OmniPT}, a model that adopts a new ``one-for-all'' paradigm that not only solves the traditional object tracking problem but also allows referring tracking based on verbal cues or semantic understanding of the tracked object.

We first addressed two issues: how to decompose the tracking task into natural language tasks that LVLMs can perform, and how to make LVLMs output results in a specified format. To this end, based on the pre-trained weights of LVLMs, we implemented a four-stage training strategy: \texttt{RL-Mid Training-SFT-RL}. We first applied the GRPO algorithm~\cite{shao2024deepseekmath} for a very lightweight reinforcement learning training to supervise the model's output format of bounding boxes. Subsequently, we used a large amount of pedestrian-related data and designed some proxy tasks. These tasks lie between the general knowledge learning in the pre-training stage and the specific tasks in the post-training stage, and like many LMs~\cite{wang2025octothinker, olmo20242}, we refer to this as the Mid Training stage. Then we performed SFT on multiple pedestrian tracking datasets. At this stage, we decoupled the tracking task into a VQA form to help the model perform the task. Finally, an additional RL stage helped the model further improve its tracking performance and instruction-following ability.

We perform the quantitative analysis of our method on several datasets including 
DanceTrack~\cite{sun2022dancetrack}, Refer-KITTI-V2~\cite{zhang2024bootstrapping},  CRTrack~\cite{chen2024cross} and BenSMOT~\cite{li2025beyond}. Most of our selected datasets are pedestrian-specific datasets, which is currently the most popular researched category. However, for RMOT, there is no pedestrian-specific dataset, so we also use these datasets for generalization tests. During test, depending on the task dataset, different instructions can be used to get an output that meets the requirements of the corresponding dataset. Experiments show that our model achieves state-of-the-art results on all datasets with a huge improvement compared to previous best methods. For example, we achieve a HOTA score of 75.04, which is a 3.06 improvement over the previous best result on BenSMOT. Additional ablation experiments also illustrate the effectiveness of the proposed method. 
\begin{figure*}[t]
  \centering
  \includegraphics[width=\linewidth]{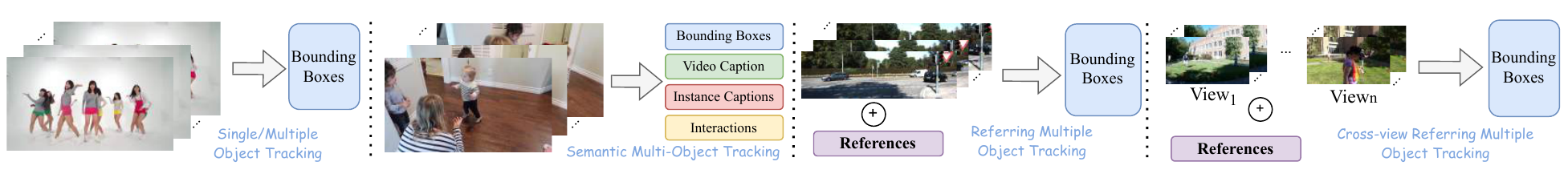}
  \caption{The format of the inputs and outputs for the different tasks our method can process.}
  \label{task}
\end{figure*}
\section{Related Work}
\label{sec:rel}

Multiple Object Tracking (MOT) aims to locate objects in a video sequence and assign an unique ID to each object across frames. There are several predominant paradigms for MOT, such as Tracking-by-Detection (TBD)~\cite{wojke2017simple,  cao2023observation, zhang2021fairmot, fu2025foundation} methods and end-to-end methods. TBD methods typically employ detectors to identify objects and associate them by computing positional or appearance-based similarities between detected objects and existing tracklets. SORT~\cite{bewley2016simple} is a pioneering approach that integrates deep learning into MOT, leveraging Kalman filtering for motion prediction and the Hungarian matching algorithm for data association. DeepSORT~\cite{wojke2017simple} further improves upon this by utilizing CNN for appearance feature extraction, combining both positional and appearance similarities for distance matrix calculation. ByteTrack~\cite{zhang2022bytetrack} refines the matching process by considering more detection results and employing a two-stage association strategy. Subsequent approaches have focused on enhancing detector performance, refining motion estimation, improving appearance feature extraction, and developing more efficient distance matrix fusion strategies.

The Transformer~\cite{vaswani2017attention} is now being widely used in MOT to form a new end-to-end paradigm~\cite{cai2022memot, zeng2022motr, zhang2023motrv2}. Trackformer~\cite{meinhardt2022trackformer} utilizes detection queries and track queries to detect new tracklets and track existing tracklets, respectively. DNMOT~\cite{fu2023denoising} adopts the ``Noising and Denoising" approach to help the model better handle objects in crowded scenarios. Several other paradigmatic approaches have also emerged. DiffMOT~\cite{lv2024diffmot} employs diffusion models to capture object motion between adjacent frames, while graph-based methods~\cite{cetintas2023unifying, li2022learning, dai2021learning, hornakova2020lifted} represent the relationships between objects as a graph structure and use graph neural networks (GNNs) to associate objects with trajectories.

\begin{figure*}[t]
  \centering
  \includegraphics[width=1\linewidth]{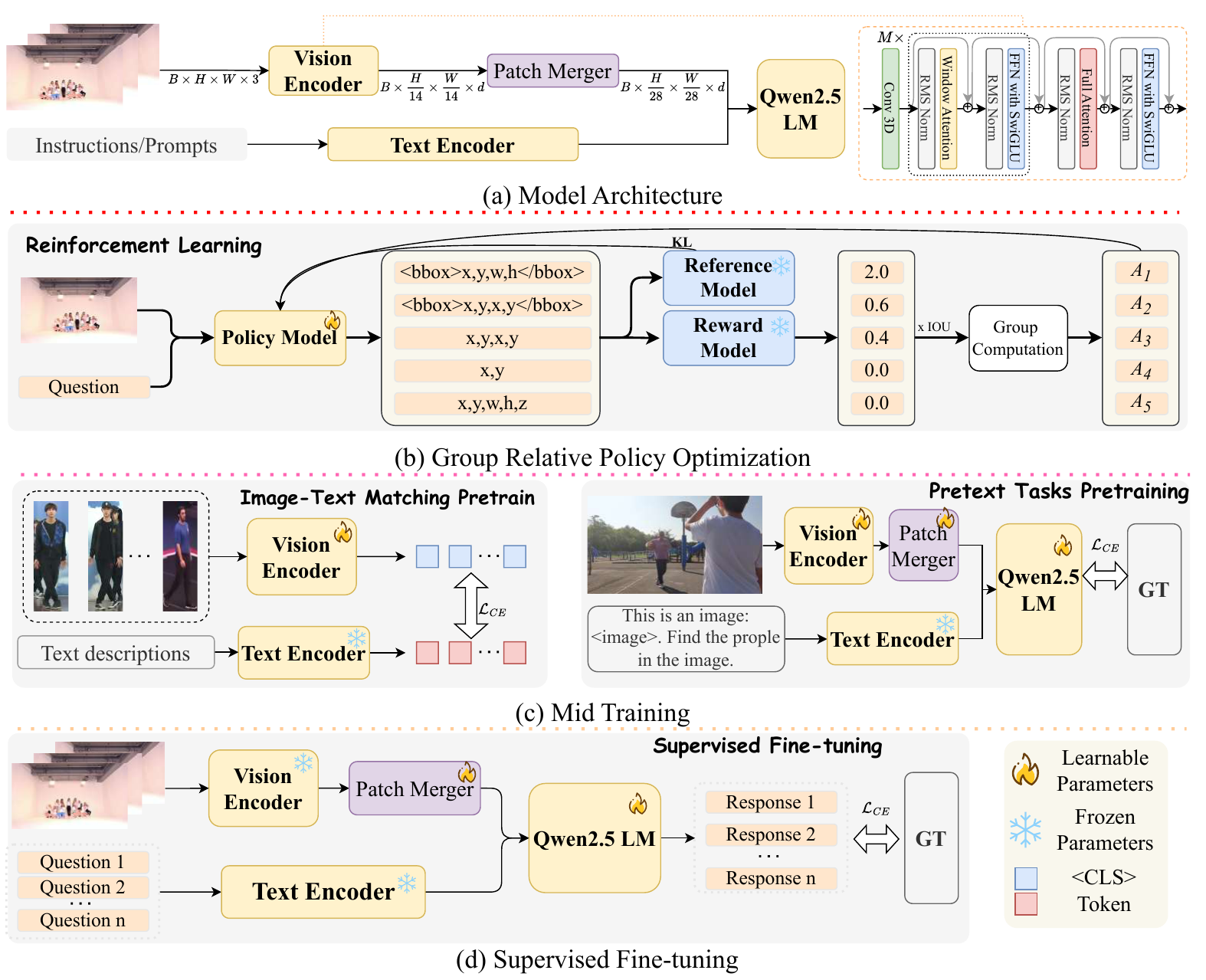}
  \caption{The model architecture of our baseline LVLM and our two-stage fine-tuning strategy for the OmniPT.}
  \label{fig3}
\end{figure*}
\subsection{Language-based Tracking Tasks}

In recent years, many new topics in the direction of object tracking have emerged, often involving multi-modality, particularly the intersection of image and natural language modalities. As shown in Figure~\ref{task}, Referring Multi-Object Tracking (RMOT)~\cite{wu2023referring, zhang2024bootstrapping} aims to track all objects in a scene that match a given linguistic description. Cross-View RMOT (CRMOT)~\cite{chen2024cross}, in contrast, performs this task across multiple viewpoints, offering a more comprehensive perspective of the objects and improving the matching between objects and their linguistic references. Semantic Multi-Object Tracking (SMOT)~\cite{li2025beyond} places greater emphasis on semantic understanding, analyzing both the scenario and the tracked objects. This task consists of four subtasks that extend beyond MOT, including video captioning, instance captioning, and categorizing interactions between different objects.

\subsection{Foundation model in MOT}
Multimodal foundation models have driven MOT research toward tracking a wider variety of objects at a finer level of granularity. SAMTrack~\cite{cheng2023segment} leverages SAM~\cite{kirillov2023segment} to segment and track all objects in a scene. OVTrack~\cite{li2023ovtrack} proposes open-vocabulary MOT, aiming to track all objects in the scene by utilizing CLIP's generalization capabilities for open-world object tracking. MASA~\cite{li2024matching} focuses on fine-grained tracking, exploring instance-level object features using SAM~\cite{kirillov2023segment} and detectors such as Grounding DINO~\cite{liu2024grounding} or YOLOX~\cite{ge2021yolox}. While these methods use LVLMs to track more objects, our approach uses language to guide the model in tracking specific objects.

\section{Methodology}
\label{sec:3}


 \subsection{Group Relative Policy Optimization}
 \label{sec:3.2}
DeepSeekMath~\cite{shao2024deepseekmath} proposed GRPO, an effective reinforcement learning algorithm. As shown in Figure~\ref{fig3}(b), it optimizes the model towards the direction of meeting ideal preferences by sampling multiple outputs and generating reward scores for these outputs. We use this step to standardize the model's output format of bounding boxes. Specifically, we want the model to use normalized coordinates and strictly follow the $\texttt{<bbox>x,y,w,h</bbox>}$ format, where $\texttt{(x,y)}$ represents the top-left coordinates and $\texttt{(w,h)}$ denotes the size of the bounding box. To this end, we adopt the following reward function:

\begin{equation}
    R(bbox_{p}) = \begin{cases} 
2 & bbox_{p} \in \texttt{set 1}, \\
0.6 & bbox_{p} \in \texttt{set 2}, \\
0.4 & bbox_{p} \in \texttt{set 3}, \\
0.0 & otherwise.
\end{cases}
\end{equation}
where $\texttt{set 1}$ means the prediction matches the format, $\texttt{set 2}$ means that the prediction contains other formats of the bounding box such as $\texttt{(x,y,x,y)}$ and $\texttt{set 3}$ means that the prediction does not contain $\texttt{<bbox>...</bbox>}$ tags. The final reward is computed as follows:
\begin{equation}
    R_{s1} = R(bbox_{p}) \times (\frac{\mathtt{IOU}(bbox_{p}, bbox_{gt})}{2}+0.5)
\end{equation}
where $\mathtt{IOU}$ is IOU score. In the first training stage, we mapped the IOU score to the range of 0.5-1, so that the model would pay more attention to the format at this stage. In the fourth training stage, we canceled this operation. We use the BenSMOT~\cite{li2025beyond} dataset as the training set in this stage and use the visual grounding as the proxy task.

 \subsection{Mid Training}
 \label{sec:3.2}

Traditional object trackers possess three primary capabilities: object detection, position prediction, and ReID. Mid training in this stage aims to enhance the model's sensitivity to language descriptions of tracked objects and improve the model's performance in all three aforementioned aspects.
As shown in Figure~\ref{fig3}(c), we first train the vision encoder of the baseline model with the image-text aligning task from CLIP~\cite{radford2021learning}. We adopt the SYNTH-PEDES~\cite{zuo2023plip} dataset for this training, a large-scale dataset for text-based person re-identification. 

Similar to CLIP, we insert special $<\!CLS\!>$ tokens into the input sequence and use them to compute similarity scores, forming a similarity matrix supervised using a cross-entropy loss function.
Furthermore, we design several pretext tasks to adapt our model to tracking-specific tasks:

\textbf{Object Detection.} For this task, we utilize all available data from the datasets. We randomly sample an image and extract the ground truth bounding boxes as the supervision.

\textbf{Location Prediction.} We sample a segment of the trajectory and provide the object’s coordinates in the first frame as input. Using instructional prompts, the model predicts the object’s position in the subsequent frame of the trajectory. The ground truth position in the next frame serves as the supervision. Notably, only the object’s position in the first frame is given, requiring the model to locate the object in the following frames before making predictions. This implicit training enhances both the model's re-identification and localization capabilities. We also include special samples where the object disappears mid-trajectory, significantly improving the model’s robustness. DanceTrack~\cite{sun2022dancetrack} is the primary dataset for this training due to its unpredictable object motion patterns.

\begin{figure*}[t]
  \centering
  \includegraphics[width=0.98\linewidth]{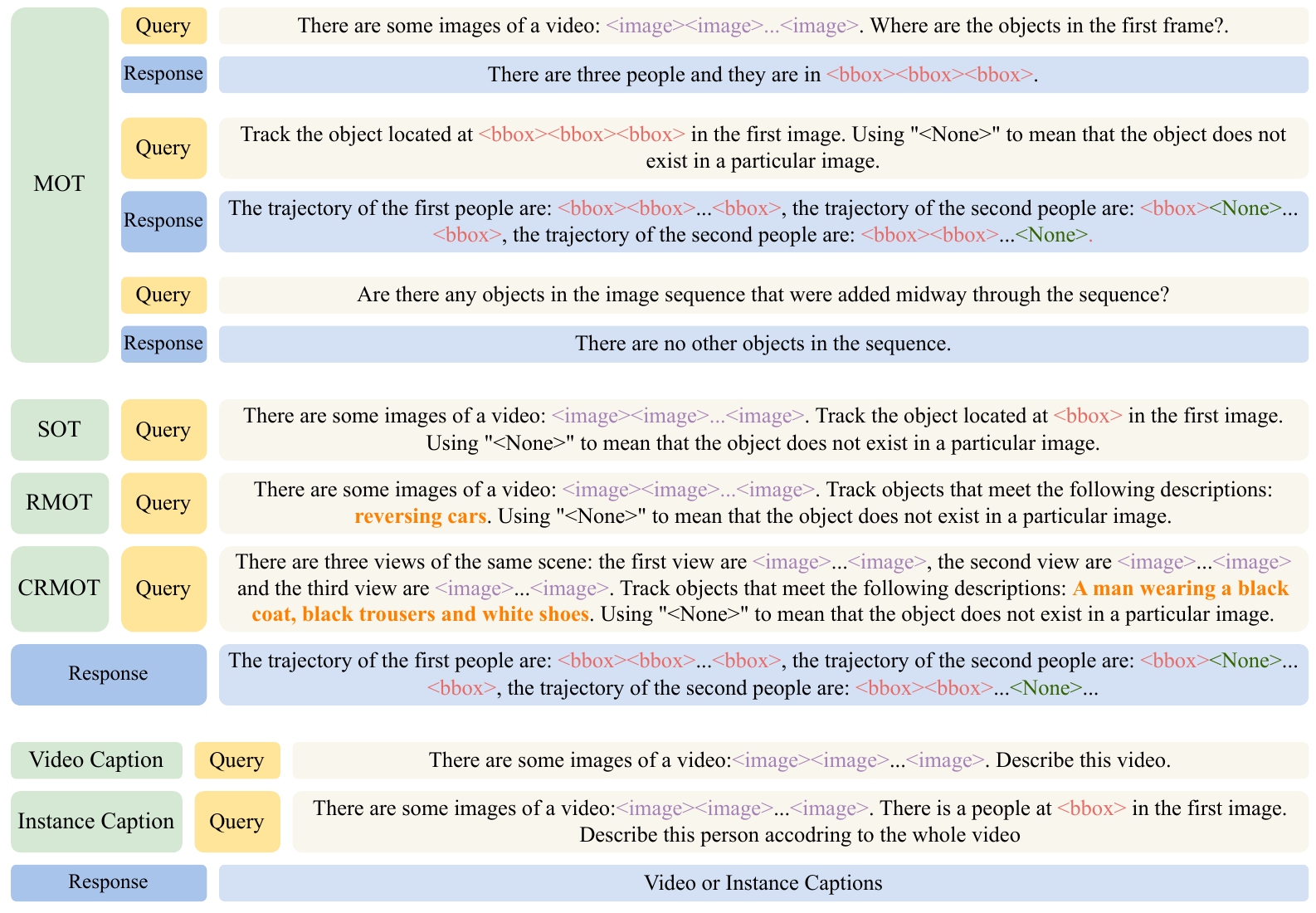}
  \caption{The format of the queries and responses for the different tasks during supervised fine-tuning. RMOT and CRMOT are based on a design for single object tracking.}
  \label{figqa}
\end{figure*}

\textbf{Person Re-identification.} We randomly sample the same person from two different frames and select other people as negative examples to construct a training sample. One frame is used as the anchor sample, and the model is guided to identify the same person from all other samples. To increase diversity, negative samples are drawn both from the same video and from different videos.

 \subsection{Supervised Fine-tuning}
\label{sec:3.3}
We conduct SFT at this stage. Based on the input and output formats of these tasks, we categorize our training samples into four primary types: multi-object tracking, referring tracking, video captioning, and instance captioning. 

\textbf{Multiple Object Tracking.} Unlike SOT, MOT requires the detection and initialization of all objects in the first frame, as well as the handling of object disappearance and the introduction of new objects throughout the sequence. Each training sample consists of a sequence of consecutive video frames and three key queries: 
\begin{itemize} 
    \item \texttt{Where are the objects located in the first frame? }
    \item \texttt{Track the location of these objects in this sequence.}
    \item \texttt{Are there any objects in the image sequence that were added midway through the sequence?}
\end{itemize} 
The model's response to the first query consists of multiple bounding boxes, presented in random order. The order of these bounding boxes is critical, as it influences the model's supervision of the second query. To ensure efficient training, we specify the true location of each object in the first frame when addressing the second query.

\textbf{Referring Multiple Object Tracking.} RMOT tracks specific objects based on references, differing from MOT by targeting a defined object class and from SOT by tracking all eligible objects rather than a single one. We replace the object position in the first frame in SOT with the linguistic description of the object.
CRMOT extends RMOT to multiple views. For each linguistic description, the task involves tracking matching objects across all views simultaneously and assigning consistent IDs to the same object in different views. To address this, we evenly divide the original sequence length among viewpoints and use natural language to explicitly specify the view associated with each sequence.

\textbf{Video Caption and Instance Caption.} These two tasks occur at the end of the object tracking pipeline, offering a semantic understanding of both the tracked object and the video. They can be viewed as  captioning tasks, where image sequences and task instructions are provided. In the instance caption task, instead of providing a cropped image of the object, the entire image is given along with the object's position in the first frame, indicating to the model which specific object in the sequence requires captioning.

\subsection{Inference}
\label{sec:3.4}
We conduct inference across four tasks: MOT, RMOT, CRMOT, and SMOT. The primary distinction between training and inference lies in the requirement for the model to track objects over extended periods during inference, whereas training involves only sequences within a training sample. To achieve long-term tracking during inference, we employ iterative multi-round dialogues. The tracking result from the last image in the sequence from the previous dialogue round serves as the initial prior information for the new round.
For SMOT, which comprises four subtasks, we exclude interaction recognition as it is not related to our research objectives. 

\begin{table*}[t]
    \centering
    \caption{Performance comparison between our method and existing methods on the BenSMOT test set. Best results are bolded.}
    \scalebox{0.7}{
    \begin{tabular}{l|cccc|cccc|ccc}
    \toprule & \multicolumn{4}{c|}{Video Caption} & \multicolumn{4}{c|}{Instance Caption} & \multicolumn{3}{c}{Tracking} \\
    \midrule    Method & BLEU $\uparrow$ & ROUGE $\uparrow$ & METEOR $\uparrow$ & CIDEr $\uparrow$ & BLEU $\uparrow$ & ROUGE $\uparrow$ & METEOR $\uparrow$ & CIDEr $\uparrow$ & HOTA$\uparrow$ & IDF1$\uparrow$ & MOTA$\uparrow$\\
    \midrule
    SORT~\cite{bewley2016simple}& 0.245 & 0.224 & 0.202 & 0.298 & 0.233 & 0.245 & 0.208 & 0.056 & 48.49 & 53.93 & 53.58\\
    DeepSORT~\cite{wojke2017simple} & 0.198 & 0.213 & 0.187 & 0.309 & 0.238 & 0.212 & 0.199 & 0.065 & 50.12 & 56.76 & 54.29\\
    OC-SORT~\cite{cao2023observation} & 0.231 & 0.252 & 0.215 & 0.242 & 0.270 & 0.205 & 0.180 & 0.033 & 51.00 & 58.01 & 55.19\\
    ByteTrack~\cite{zhang2022bytetrack} & 0.224 & 0.225 & 0.212 & 0.266 & 0.304 & 0.242 & 0.224 & 0.064 & 68.84 & 78.37 & 73.87\\
    TransTrack~\cite{sun2020transtrack} & 0.247 & 0.248 & 0.209 & 0.269 & 0.283 & 0.219 & 0.201 & 0.074 & 71.31 & 78.67 & 74.08\\
    MOTR~\cite{zeng2022motr} & 0.187 & 0.254 & 0.203 & 0.244 & 0.230 & 0.209 & 0.182 & 0.061 & 66.10 & 68.97 & 45.19\\
    MOTRv2~\cite{zhang2023motrv2} & 0.217 & 0.258 & 0.219 & 0.248 & 0.238 & 0.241 & 0.204 & 0.059 & 65.28 & 70.76 & 45.52\\
    SMOTer~\cite{li2025beyond} & 0.245 & 0.261 & 0.223 & 0.343 & 0.306 & 0.223 & 0.209 & 0.087 & 71.98 & 80.65 & 77.71\\
    \midrule
    OmniPT (Ours)& \textbf{0.519} & \textbf{0.488} & \textbf{0.459} & \textbf{1.826} & \textbf{0.512} & \textbf{0.342} & \textbf{0.353} & \textbf{0.482} & \textbf{75.04} & \textbf{81.13} & \textbf{77.78}\\
    \bottomrule
    \end{tabular}
    }
    \label{tab1}
\end{table*}

\begin{table}[t]
    \centering
    \caption{Performance comparison between our method and existing methods on the CRMOT test set.}
    \scalebox{0.58}{
    \begin{tabular}{l|cc|cc}
    \toprule & \multicolumn{2}{c|}{CRMOT In-domain}  &\multicolumn{2}{c}{CRMOT Cross-domain} \\
    \midrule    Method & CVRIDF1 $\uparrow$ & CVRMA $\uparrow$& CVRIDF1 $\uparrow$ & CVRMA $\uparrow$  \\
    \midrule
    TransRMOT~\cite{wu2023referring} & 23.30 & 8.03 & 3.66 & 0.2 \\
    TempRMOT~\cite{zhang2024bootstrapping} & 23.43 & 10.14 & 3.78& 0.39\\
    CRTracker~\cite{chen2024cross} & 54.88 & 35.97&12.52&2.32 \\
    \midrule
    OmniPT (Ours)& \textbf{62.13}&\textbf{42.39} & \textbf{46.54} & \textbf{33.67}\\
    \bottomrule
    \end{tabular}
    }
    \label{tab2}
\end{table}

\begin{table}[t]
        \caption{Performance comparison between our method and existing methods on the DanceTrack test set.}
    \centering
    \scalebox{0.65}{
    \begin{tabular}{l|ccccc}
    \toprule    Method & HOTA $\uparrow$ & DetA$\uparrow$ & AssA$\uparrow$ &MOTA $\uparrow$ & IDF1 $\uparrow$ \\
    \midrule
    FairMOT~\cite{zhang2021fairmot} & 39.7 & 66.7 & 23.8 & 82.2 & 40.8 \\
    TransTrack~\cite{sun2020transtrack} & 45.5 & 75.9 & 27.5 & 88.5 & 45.2\\
    MOTR~\cite{zeng2022motr} &54.2 & 73.5 & 40.2 & 79.7 & 51.5 \\
    ByteTrack~\cite{zhang2022bytetrack} & 47.7  & 71.0 & 32.1 & 89.6 & 53.9\\
    OC-SORT~\cite{cao2023observation} & 55.1 & 80.3 & 38.3 & \textbf{92.0} & 54.6\\
    \midrule
    OmniPT (Ours) & \textbf{56.4}& \textbf{81.7}& \textbf{41.0} & 90.2&\textbf{55.4}\\
    \bottomrule
    \end{tabular}
    }
    \label{tab3}
\end{table}

\begin{table}[t]
        \caption{Performance comparison between our method and existing methods on the Refer-KITTI-v2 test set.}
    \centering
    \scalebox{0.66}{
    \begin{tabular}{l|ccccc}
    \toprule    Method & HOTA $\uparrow$ & DetA$\uparrow$ & AssA$\uparrow$ & MOTA $\uparrow$ & IDF1 $\uparrow$ \\
    \midrule
    FairMOT~\cite{zhang2021fairmot} & 22.53 & 15.8 & 32.82 & - & - \\
    ByteTrack~\cite{zhang2022bytetrack} & 24.59  & 16.78 & 36.63 & - & -\\
    iKUN~\cite{du2024ikun} & 10.32 & 2.17 & 49.77 & - & -\\
    TransRMOT~\cite{wu2023referring} & 31.00 & 19.40 & 49.68 & - &-\\
    TempRMOT~\cite{zhang2024bootstrapping} & 35.04 & 22.97 & 53.58& - & - \\
    \midrule
    OmniPT (Ours) & \textbf{36.15} & \textbf{26.68} &  \textbf{54.62} & \textbf{76.56} & \textbf{65.30} \\
    \bottomrule
    \end{tabular}
    }
    \label{tab4}
\end{table}

\section{Experiments}
\label{4_experiments}

\subsection{Datasets and Metrics}

\textbf{MOT.} For MOT evaluation, we select DanceTrack~\cite{sun2022dancetrack} as our primary benchmark among numerous datasets such as MOT17~\cite{milan2016mot16}, MOT20~\cite{dendorfer2020mot20} and SportsMOT~\cite{cui2023sportsmot}. The MOT Challenge series, particularly MOT20, often contains densely populated scenes with over 200 people simultaneously, which poses a challenge for our approach due to hardware limitations on the maximum output length in input tokens. While SportsMOT is a large dataset, it lacks comprehensive labeling in certain scenarios (\textit{e.g.}, labeling only players but ignoring spectators on a basketball court), making it less suitable for our purposes.
DanceTrack features highly unpredictable object movements and significant appearance similarity between objects, making it an ideal choice for evaluating MOT performance.

\textbf{SMOT.} We evaluate our method using BenSMOT~\cite{li2025beyond}, a dataset comprising 3,292 videos. For each video, we perform tracking on every object and generate captions for both the video as a whole and each individual person within it.

\textbf{RMOT.} We evaluate our model's Referring MOT capability using Refer-KITTI-V2~\cite{zhang2024bootstrapping}. Compared to its predecessor, Refer-KITTI-V2 expands the data size. Notably, this dataset is not limited to pedestrian tracking but also includes a wide range of objects, such as cars. While our primary focus is pedestrian tracking, this dataset allows us to evaluate the generalization performance of our model across diverse object categories.

\textbf{CRMOT.} We evaluate the CRMOT capabilities of our model using CRTrack~\cite{chen2024cross}, a dataset comprising 13 scenarios. Each scenario is divided into three to four perspectives and includes corresponding linguistic descriptions for the tracked objects.

\textbf{Metrics.} For the MOT, RMOT, and tracking tasks in SMOT, we employ HOTA~\cite{luiten2021hota} and CLEAR~\cite{bernardin2008evaluating} as evaluation metrics. For the caption tasks, we adopt ROUGE-L~\cite{lin2004rouge}, BLEU~\cite{papineni2002bleu}, METEOR~\cite{banerjee2005meteor}, and CIDEr-D~\cite{vedantam2015cider} to assess performance. For the CRMOT task, we introduce CVRIDF1 and CVRMA~\cite{chen2024cross} as our evaluation metrics.

\subsection{Implementation Details}
The proposed method is implemented using PyTorch and trained on 24 NVIDIA A100 GPUs. We apply LoRA-based~\cite{hu2021lora} fine-tuning to the parameters, as illustrated in Figure~\ref{fig3}. Depending on the number of training samples, we set the training epochs from 3 to 5. The maximum number of image pixels is set to $28\times 28 \times 646$, approximating a $1920\times 1080$ image downsampled twice. The maximum output length of 2048. We use a $\texttt{warm up-constant-decay}$ learning rate strategy with a peak learning rate of 1e-5. During training, we process 16 images per sample (increased to 32 for semantic understanding tasks), randomly sampling images for training and processing video frames sequentially during inference.

\begin{table*}[t]
    \centering
    \caption{Performance comparison between different LVLMs and between different model scale.}
    \scalebox{0.85}{
    \begin{tabular}{l|c|cccc|ccc|c}
    \toprule \multirow{2}{*}{LVLM}& \multirow{2}{*}{Scale}& \multicolumn{4}{c|}{Instance Caption} & \multicolumn{3}{c|}{Tracking}&RMOT \\
    \cmidrule{3-10}
         & & BLEU $\uparrow$ & ROUGE $\uparrow$ & METEOR $\uparrow$ & CIDEr $\uparrow$ & HOTA$\uparrow$ & IDF1$\uparrow$ & MOTA$\uparrow$& HOTA$\uparrow$\\
    \midrule
    LLaVA-NeXT~\cite{li2024llava}&8B &0.211 & 0.149&0.172 &0.244 & 49.7& 48.6& 80.4&26.73\\
    InternVL2.5~\cite{chen2024expanding}&4B &0.186 & 0.173& 0.217& 0.017&46.5 &48.2& 77.4&25.02\\
    Qwen2-VL~\cite{wang2024qwen2}&2B &0.370& 0.303& 0.251& 0.256 & 53.4 & 51.1& 84.3&33.37\\
    \midrule   
    Qwen2.5-VL~\cite{bai2025qwen2}&3B & 0.379 & 0.314& 0.244& 0.263& 53.8& 51.2& 87.4& 34.70\\
    Qwen2.5-VL~\cite{bai2025qwen2}&7B & 0.446 & 0.323& 0.278& 0.316&54.4 & 52.3& 90.1&34.92\\
    Qwen2.5-VL~\cite{bai2025qwen2}&72B  & \textbf{0.512} &\textbf{0.342} &\textbf{0.353} & \textbf{0.482}& \textbf{56.4} & \textbf{55.4} & \textbf{90.2} &\textbf{36.15}\\
    \bottomrule
    \end{tabular}
    }
    \label{tab5}
\end{table*}

\subsection{Main Results}
\textbf{Semantic understanding.} Table~\ref{tab1} presents the results of our approach for semantic understanding tasks on BenSMOT, including the video caption and instance caption tasks. The results of the baseline methods are sourced from the SMOTer~\cite{li2025beyond}. The experimental results highlight the robustness of our approach, particularly leveraging the strength of LVLM in captioning tasks. Our method achieves improvements of over 50\% across all metrics, with the CIDEr metric showing a remarkable fivefold increase.

\textbf{Multi-object tracking.} 
For the MOT task, Tables~\ref{tab1} and ~\ref{tab3} highlight the strong performance of our method. We achieve 75.04 and 56.4 HOTA results on the BenSMOT and DanceTrack datasets, respectively. Notably, on DanceTrack, our method leverages robust location prediction and person re-identification capabilities to effectively distinguish between similar objects, resulting in superior tracking performance.

\textbf{Referring MOT.} Table~\ref{tab4} presents our results on Refer-KITTI-v2. Our approach achieves state-of-the-art performance on the HOTA metric, with significant improvements of 26.68 on DetA and 54.62 on AssA, underscoring the effectiveness of our proposed method. Additionally, we achieve 75.56 MOTA and 65.30 IDF1, further validating the robustness of our approach.

\textbf{Cross-view Referring MOT.} CRTrack integrates two  datasets and is divided into In-domain and Cross-domain setups based on whether the test video originates from the same dataset as the training videos. All methods perform significantly better in the In-domain setting due to the absence of a domain gap between the training and test sets. Our method achieves a CVRIDF1 of 62.13 and a CVRMA of 42.39 in the In-domain setting, and a CVRIDF1 of 46.54 and a CVRMA of 33.67 in the another setting, demonstrating the effectiveness of our approach in both settings. 

\subsection{Ablation Studies}

\begin{table}[t]
    \centering
    \caption{Effect of different stage of training on the final results. MT represent the Mid Training and RL represent the fourth stage of reinforcement learning. We use the HOTA, HOTA, and CIDEr as the metrics for each task.}
    \scalebox{1}{
    \begin{tabular}{ccc|ccc}
    \toprule \multicolumn{3}{c|}{Training Stages} & \multicolumn{3}{c}{Tasks}  \\
    \midrule   MT& SFT&RL  & MOT & RMOT & SMOT\\
    \midrule
    &\checkmark& &   47.38 & 30.37 & 0.40\\
    \checkmark&\checkmark& &  51.89& 33.26& 0.44\\ 
    & \checkmark & \checkmark & 48.63& 35.46 & 0.41\\
    \checkmark &\checkmark & \checkmark  & \textbf{56.40}& \textbf{36.15} & \textbf{0.48}\\
    \bottomrule
    \end{tabular}
    }
    \label{tab6}
\end{table}
\textbf{LVLMs and model scale.} We show in Table~\ref{tab5} the effect of different LVLMs and different scales. We choose to use LLaVA-NeXT~\cite{li2024llava}, Qwen2 VL~\cite{wang2024qwen2} and InternVL2.5~\cite{chen2024expanding} as the comparison methods of our method. We also chose different scales of the QwenVL-2.5 model for the ablation experiments. The Qwen2.5-VL achieved the best results, and we believe that it is the design of the dynamic resolution that leads to a model that can capture the semantic features of the object more clearly, which facilitates the performance of every task.
\begin{table}[t]
    \centering
    \caption{Effect of the number of images. N/A represents the inability of the model to handle long inputs. }
    \scalebox{0.76}{
    \begin{tabular}{c|cc|cc|cc}
    \toprule \multirow{2}{*}{Length}& \multicolumn{2}{c|}{Video Caption} & \multicolumn{2}{c|}{Instance Caption}&  \multicolumn{2}{c}{Tracking}\\
    \cmidrule{2-7}
       & METEOR&CIDEr  & METEOR & CIDEr & HOTA & MOTA\\
    \midrule
    2 & 0.197& 0.773 & 0.149 & 0.203& 52.19 & 72.36\\
    4 & 0.258 & 1.027 & 0.199 & 0.271& 70.60 & 76.42\\
    8 & 0.344 & 1.370 & 0.265 & 0.362& \textbf{73.04} & \textbf{77.78}\\
    16 & 0.418 & 1.774& 0.336& 0.457& N/A & N/A \\
    32 & \textbf{0.459} & \textbf{1.826} & \textbf{0.353}& \textbf{0.482}& N/A&N/A \\
    \bottomrule
    \end{tabular}
    }
    \label{tab7}
\end{table}

\textbf{The different stages of the fine-tuning.} We investigate the impact of different stages of fine-tuning in Table~\ref{tab6}. Stage 1 is necessary, otherwise we would not be able to monitor the wide variety of outputs in a uniform manner. The results show that gradually increasing pre-training tasks enhance model performance. The first row represents our baseline results \textbf{without additional data} to ensure fairness. Mid training stage significantly improves the model's ability to capture pedestrian details, leading to a substantial improvement in tracking performance. RL can further improve the model performance after sufficient training, however, the improvement is very limited when it is not sufficiently trained.

\textbf{Image length in the training samples.} In each training iteration, we provide the model with a set of images for tracking or semantic understanding. Using BenSMOT~\cite{li2025beyond}, we explore the effect of the number of images per iteration on performance. As shown in Table~\ref{tab7}, increasing the number of images enhances the model's understanding of objects and scenes. However, for tracking tasks, a smaller number of images leads to more inference rounds, which can weaken the model's memory of the object. Conversely, a larger number of images challenges the model's ability to process multiple images simultaneously. As a result, the overall trend for HOTA and MOTA metrics initially rises and then declines until it cannot be converged.

\subsection{Limitation} 
On the one hand, LVLMs struggle to accurately track or even locate all objects in scenes with a large number of objects, such as those in the MOT20~\cite{dendorfer2020mot20} dataset. This limitation highlights a critical direction for future advancements in LVLM capabilities. On the other hand, while our work focuses on the “pedestrians” category, we aim to inspire research toward developing unified open-vocabulary object trackers capable of handling diverse object categories.

\section{Conclusion}
We propose OmniPT, a novel tracker capable of leveraging object semantic information to perform four tasks simultaneously: MOT, RMOT, SMOT and CRMOT, guided by instructions. Our model is trained through a four-stage training processes on an LVLM baseline. In the first stage, we employed reinforcement learning to standardize the model's output format. Subsequently, we utilized mid-training and supervised fine-tuning to enable the model to learn how to use semantic information for continuous object tracking. Finally, we further enhanced the model's tracking performance and instruction-following ability through reinforcement learning. Extensive experiments on several benchmarks demonstrate our state-of-the-art performance, and ablation studies validate the effectiveness of our approach.

\section{Acknowledgments}
This work was supported in part by the National Key R\&D Program of China (No.2021ZD0112803), the National
Natural Science Foundation of China (No.62176060), and the Program for Professor of Special Appointment (Eastern Scholar) at Shanghai Institutions of Higher Learning.

\bigskip

\bibliography{aaai2026}

\end{document}